\documentclass[11pt, a4paper]{article}

\usepackage[utf8]{inputenc}
\usepackage{geometry}
\usepackage{hyperref}
\usepackage{booktabs} 
\usepackage{listings} 
\usepackage{xcolor}
\usepackage{url}
\usepackage{caption}  
\usepackage{rotating} 
\usepackage{tabularx} 
\usepackage{pdflscape}
\usepackage{float}
\usepackage{array}
\usepackage{natbib}
\usepackage{amsmath}
\usepackage{amssymb}
\usepackage{siunitx} 
\usepackage{threeparttable}
\usepackage{colortbl} 

\hypersetup{
    colorlinks=true,
    linkcolor=blue,
    filecolor=magenta,      
    urlcolor=cyan,
    citecolor=blue
}

\title{\textbf{Heterogeneous Connectivity in Sparse Networks: Fan-in Profiles, Gradient Hierarchy, and Topological Equilibria}}
\author{Nikodem Tomczak \\ nikodem.tomczak@gmail.com \\ \textit{Thulge Labs, Singapore}}
\date{November 2025}

\begin{document}

\maketitle

\begin{abstract}
Profiled Sparse Networks (PSN) replace uniform connectivity with deterministic, heterogeneous fan-in profiles defined by continuous, nonlinear functions, creating neurons with both dense and sparse receptive fields. We benchmark PSN across four classification datasets spanning vision and tabular domains, input dimensions from 54 to 784, and network depths of 2--3 hidden layers. At 90\% sparsity, all static profiles, including the uniform random baseline, achieve accuracy within 0.2--0.6\% of dense baselines on every dataset, demonstrating that heterogeneous connectivity provides no accuracy advantage when hub placement is arbitrary rather than task-aligned. This result holds across sparsity levels (80--99.9\%), profile shapes (eight parametric families, lognormal, and power-law), and fan-in coefficients of variation from 0 to 2.5. Internal gradient analysis reveals that structured profiles create a 2--5$\times$ gradient concentration at hub neurons compared to the ${\sim}1\times$ uniform distribution in random baselines, with the hierarchy strength predicted by fan-in coefficient of variation ($r = 0.93$). When PSN fan-in distributions are used to initialise RigL dynamic sparse training, lognormal profiles matched to the equilibrium fan-in distribution consistently outperform standard ERK initialisation, with advantages growing on harder tasks, achieving +0.16\% on Fashion-MNIST ($p = 0.036$, $d = 1.07$), +0.43\% on EMNIST, and +0.49\% on Forest Cover. RigL converges to a characteristic fan-in distribution regardless of initialisation. Starting at this equilibrium allows the optimiser to refine weights rather than rearrange topology. Which neurons become hubs matters more than the degree of connectivity variance, i.e., random hub placement provides no advantage, while optimisation-driven placement does.
\end{abstract}

\paragraph{Key Contributions.}
\begin{enumerate}
\item Deterministic nonlinear fan-in profiles for structured heterogeneous sparsity, enabling continuous parameterisation of connectivity distributions as an architectural variable.
\item Practical derivation and empirical validation of mean fan-in initialisation for heterogeneous sparse networks, with immediate relevance to any method that produces non-uniform connectivity.
\item An experimental framework that decouples capacity distribution (which neurons receive how many connections) from input coverage (which specific inputs connect to each neuron).
\item Across four datasets, sparsity levels from 80\% to 99.9\%, and fan-in CVs from 0 to 2.5, static connectivity structure does not significantly affect accuracy at matched parameter counts.
\item Fan-in CV determines gradient heterogeneity as a structural consequence of mask geometry ($r = 0.93$), independently of profile shape or task performance.
\item Lognormal initialisation matched to the RigL equilibrium fan-in distribution consistently outperforms ERK, with advantages increasing on harder tasks.
\end{enumerate}

\section{Introduction}
Deep neural networks achieve remarkable performance across diverse domains, yet the vast majority of parameters contribute minimally to final predictions.\citep{Frankle2019, Han2015} Sparse neural networks address this inefficiency by eliminating redundant connections while maintaining competitive accuracy.\citep{Hoefler2021,Gale2019} Early approaches to sparsity remove individual weights based on magnitude or saliency, treating each connection independently without considering the connectivity structure of the neurons they connect.\citep{LeCun1990, Han2015} This treatment ignores organizational principles observed in complex network systems, where connectivity patterns exhibit structured heterogeneity with hub nodes maintaining dense connectivity while peripheral nodes maintain sparse, selective connections.\citep{Barabasi1999, Bullmore2009} Recent empirical evidence from dynamic sparse training methods provides support for the importance of heterogeneous connectivity. At high sparsity levels gradient-based rewiring evolves toward connectivity patterns with high variance in per-neuron fan-in, despite starting from uniform random initialization.\citep{lasby2024dynamic} These findings suggest that networks seek heterogeneous connectivity organization when given the flexibility to evolve their topology, yet standard sparse training methods ignore this preference by imposing uniform constraints from initialization. Dynamic topology search is computationally expensive, requiring gradient computation for potential connections, periodic topology updates involving global weight ranking and selection, and storage of metadata tracking connection presence across updates. If dynamic sparse training methods naturally evolve toward heterogeneous connectivity patterns, can we accelerate convergence and improve efficiency by designing beneficial structure from initialization? If the target structure is predictable, as suggested by the consistency of emergent patterns across different architectures and sparsity levels, then starting with appropriate structure might bypass costly search while preserving its benefits.

We introduce Profiled Sparse Networks (PSN), an architecture that designs heterogeneous connectivity from initialization rather than waiting for it to emerge through gradient-based rewiring. PSN assigns per-neuron fan-in according to continuous nonlinear profile functions that map neuron index to connectivity density, creating deterministic spatial arrangements of hub and specialist neurons within individual layers. By parameterizing connectivity distributions through interpretable mathematical functions, PSN enables exploration of the connectivity design space as controlled experimental variables. This approach contrasts with both uniform random sparsity, which treats all neurons identically, and dynamic sparse training, which discovers structure through topology search. The research question we address is whether structured heterogeneous connectivity provides superior inductive bias compared to uniform random sparsity when parameter counts are held constant. Recent concurrent work has begun exploring designed heterogeneous topologies from complementary angles. The Dendritic Network Model (DNM)\citep{Cerretti2025dendritic} generates sparse initializations through parametric distributions of dendrites, receptive fields, and degree, enabling control over modularity and degree heterogeneity. Brain-inspired topological models from the same group\citep{zhang2026brain} demonstrate that scale-free and small-world initializations can outperform random baselines at extreme sparsity, with extensions to transformers and large language models.

We evaluate PSN on four classification benchmarks including MNIST\citep{LeCun1998}, Fashion-MNIST\citep{Xiao2017}, EMNIST-Balanced\citep{Cohen2017}, and Forest Cover Type\citep{Blackard1999}, spanning input dimensions from 54 to 784 and network depths of 2--3 hidden layers. We derive the initialisation requirements for heterogeneous sparse networks from forward and backward variance analysis and validate them empirically. A multi-peak interpolation experiment confirms that fan-in CV alone predicts gradient concentration ($r = 0.93$), independent of profile shape. When lognormal profiles are used to initialise RigL, matching the equilibrium fan-in distribution outperforms ERK with advantages growing on harder tasks. Static connectivity structure does not affect accuracy at matched sparsity, but understanding the structural endpoints of dynamic sparse training provides a principled basis for initialisation design.

\section{Methods and the PSN Framework}
\subsection{Datasets and Network Architectures}

We evaluate PSN across four datasets spanning vision and tabular domains (Table~\ref{tab:datasets}). All image datasets use flattened $28 \times 28$ grayscale inputs (dimension 784). Forest Cover uses 54 tabular features.

\begin{table}[tbp]
    \centering
    \caption{Datasets used in this study.}
    \label{tab:datasets}
    \small
    \renewcommand{\arraystretch}{1.2}
    \begin{tabularx}{0.90\textwidth}{l >{\centering\arraybackslash}X >{\centering\arraybackslash}X >{\centering\arraybackslash}X >{\centering\arraybackslash}X}
        \toprule
        \textbf{Dataset} & \textbf{Train} & \textbf{Test} & \textbf{Input dim} & \textbf{Classes} \\
        \midrule
        MNIST\citep{LeCun1998}                     & 60{,}000  & 10{,}000 & 784 & 10 \\
        Fashion-MNIST\citep{Xiao2017}              & 60{,}000  & 10{,}000 & 784 & 10 \\
        EMNIST-Balanced\citep{Cohen2017}           & 112{,}800 & 18{,}800 & 784 & 47 \\
        Forest Cover\citep{Blackard1999}           & 464{,}809 & 116{,}203 & 54 & 7  \\
        \bottomrule
    \end{tabularx}
\end{table}

All experiments use multi-layer perceptrons (MLPs) with LayerNorm and ReLU activations. The three 784-input datasets share a common 784$\to$1024$\to$1024 hidden architecture with 10 or 47 output neurons. Forest Cover uses a deeper 54$\to$1024$\to$1024$\to$1024$\to$7 architecture (three hidden layers), selected by grid search over widths and depths . PSN sparse masks are applied to all hidden layers; the output layer is kept dense. Output layers are small relative to hidden layers and contribute negligibly to total parameter count. Sparsifying them provides little compression benefit while degrading class separability. Table~\ref{tab:architectures} summarises the training configurations.

\paragraph{Forest Cover architecture selection.}\label{sec:forest-arch}
We selected the Forest Cover architecture by grid search over hidden widths $\{256, 512, 1024, 2048\}$ and depths $\{2, 3\}$ hidden layers, training dense models for 20 epochs across 3 seeds. Adding a third hidden layer improved accuracy by 1.3--1.5\% at every width. We selected 3 hidden layers of width 1024 (95.07\%~$\pm$~0.04\% test accuracy), which matches the MNIST hidden width for direct comparison.

\begin{table}[tbp]
    \centering
    \caption{Network architectures and training configurations.}
    \label{tab:architectures}
    \small
    \renewcommand{\arraystretch}{1.2}
    \begin{tabularx}{0.90\textwidth}{l >{\centering\arraybackslash}X >{\centering\arraybackslash}X}
        \toprule
        \textbf{Dataset} & \textbf{Architecture} & \textbf{Batch size} \\
        \midrule
        MNIST              & $784\!\to\!1024\!\to\!1024\!\to\!10$  & 128 \\
        Fashion-MNIST      & $784\!\to\!1024\!\to\!1024\!\to\!10$  & 128 \\
        EMNIST-Balanced    & $784\!\to\!1024\!\to\!1024\!\to\!47$  & 128 \\
        Forest Cover       & $54\!\to\!1024\!\to\!1024\!\to\!1024\!\to\!7$ & 256 \\
        \bottomrule
    \end{tabularx}
\end{table}

\subsection{Connectivity Profiles and Structured Heterogeneity}

PSN introduces structured heterogeneity through connectivity profile functions that deterministically map neuron indices to fan-in (number of incoming connections). In a fully connected layer $x \in \mathbb{R}^n \to y \in \mathbb{R}^m$, dense networks assign uniform fan-in $f_i = n$ and random sparse networks assign expected fan-in $\mathbb{E}[f_i] = (1-s)n$ identically. PSN instead assigns fan-in deterministically via a profile function $P : [0,1] \to [0,1]$ evaluated at the normalized neuron index $t = i/m$, inducing heterogeneous capacity allocation within a layer. All profiles preserve the same mean fan-in $\bar{f} = n(1-s)$; they differ only in the connectivity coefficient of variation (CCV = $\sigma_f / \mu_f$).

\begin{figure}[H]
    \centering
    \includegraphics[width=\textwidth]{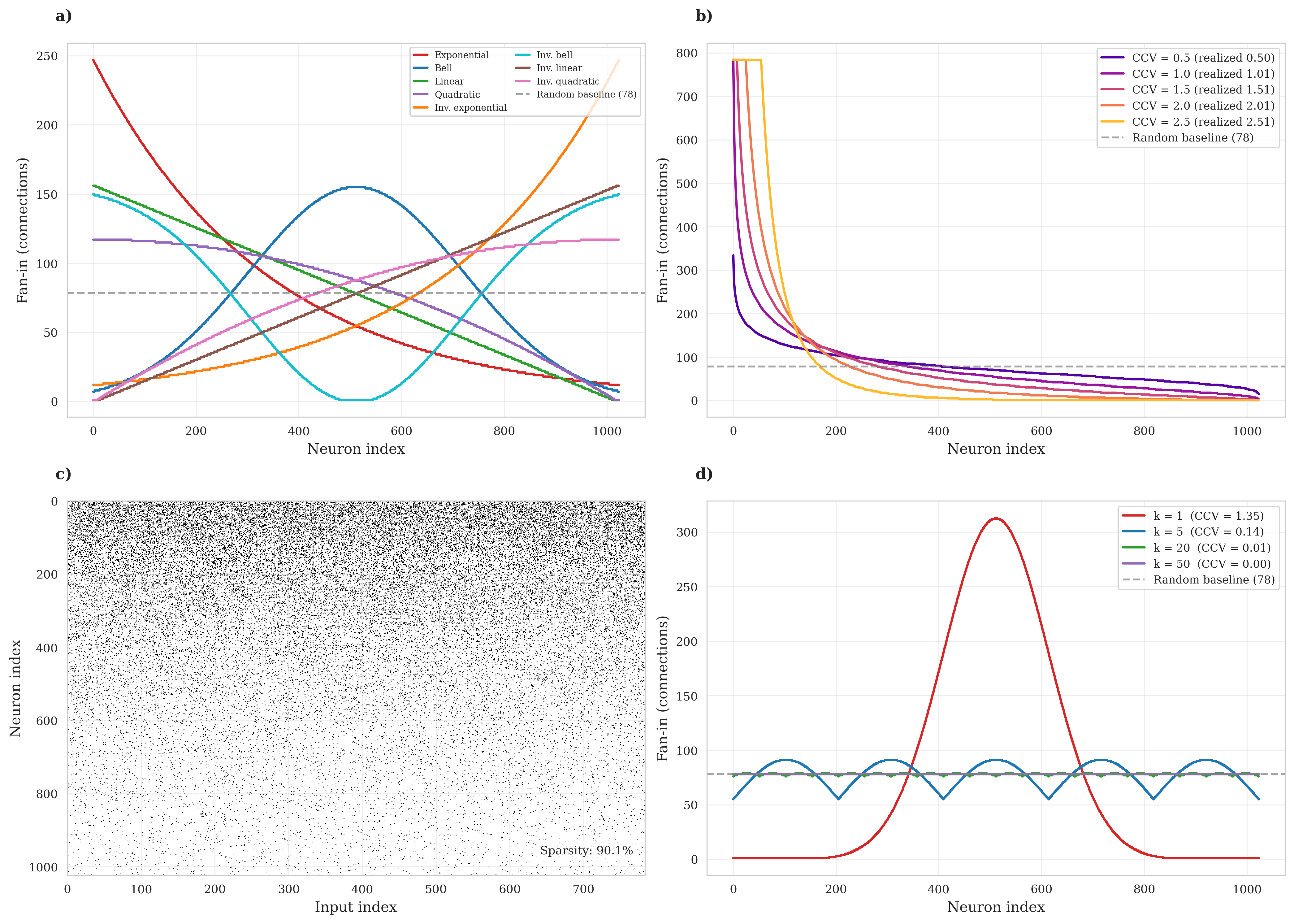}
    \caption{PSN methods overview at 90\% sparsity for a 784$\to$1024 layer. \textbf{(a)} Eight parametric fan-in profiles. The exponential profile assigns hub neurons $>$250 connections while the quadratic profile reaches $\sim$120. Inverse profiles mirror the spatial arrangement of hubs and specialists. The dashed line marks the random baseline (uniform fan-in 78). \textbf{(b)} Lognormal profiles at five target CCVs (0.5--2.5). Fan-in values are sampled from lognormal distributions, sorted in descending order, and assigned to neuron indices. Higher CCV produces a heavier tail. \textbf{(c)} Binary connectivity mask for the exponential profile with random spreading. Each row corresponds to one output neuron, each column to one input. Connection density decreases from top (high fan-in) to bottom (low fan-in). \textbf{(d)} Multi-peak profile interpolation at $\beta = 50$. Increasing peak count $k$ from 1 to 50 smoothly transitions from extreme heterogeneity (CCV = 1.23) to near-uniformity (CCV = 0.00). All profiles share the same global sparsity and mean fan-in.}
    \label{fig:fan-in-profiles}
\end{figure}

We define eight parametric profiles that control how a layer's connection budget is allocated across neurons (Figure~\ref{fig:fan-in-profiles}a). Monotonic profiles include:
\begin{align}
P_{\mathrm{lin}}(t) &= 1 - t, \\
P_{\mathrm{quad}}(t) &= 1 - t^2, \\
P_{\mathrm{exp}}(t) &= \exp(-\alpha t),
\end{align}
with $\alpha = 3$. These profiles interpolate from dense hub neurons (low indices) to sparse specialist neurons (high indices). Inverse variants swap hub and specialist roles: $P_{\mathrm{inv}}(t) = P(1-t)$. We also consider a non-monotonic bell-shaped profile:
\begin{equation}
P_{\mathrm{bell}}(t) = \exp\!\left(-\beta (2(t - 0.5))^2\right),
\end{equation}
with $\beta = 3$, concentrating connectivity in the middle of the layer. Its complement $P_{\mathrm{inv\_bell}}(t) = 1 - P_{\mathrm{bell}}(t)$ places high connectivity at the layer boundaries.

A complementary profile family is defined by sampling fan-in values from a lognormal distribution. Given a target mean fan-in $\bar{f} = n(1-s)$ and a target CCV, we set the lognormal shape parameters:
\begin{equation}
\sigma^2 = \ln(1 + \mathrm{CCV}^2), \qquad \mu = \ln \bar{f} - \tfrac{1}{2}\sigma^2.
\end{equation}
Fan-in values are drawn from $\mathrm{LogNormal}(\mu, \sigma^2)$, clamped to $[f_{\min},\, n]$, and sorted in descending order, assigning the highest fan-in to neuron index 0 (Figure~\ref{fig:fan-in-profiles}b). Because clamping compresses the tails, the realized CCV is lower than the theoretical value; we therefore use binary search over $\sigma$ to find the shape parameter that achieves the target CCV after clamping. The lognormal family is parameterized directly by CCV, enabling precise control over fan-in heterogeneity. We use lognormal profiles as initialisation for dynamic sparse training.

A power-law profile assigns fan-in proportional to neuron rank (rank-frequency law), with the rank exponent chosen to match a target CCV. Biological neural circuits exhibit Pareto exponents $\approx 2$--$3$ (CCV $\approx 0.7$--$1.4$). The power-law profile achieves lower maximum CCV than lognormal at the same sparsity ($\sim$2.1 vs.\ $\sim$2.8 at 90\%) and is included as a biologically motivated reference.

\subsection{Sparsity Matching and Fan-in Constraints}

Profile values specify relative connectivity but not absolute sparsity. For a target sparsity $s$, we define fan-in for neuron $i$ as
\begin{equation}
f_i = \max\!\left(f_{\min},\; \left\lceil \lambda \, P\!\left(\tfrac{i}{m}\right) \right\rceil \right),
\end{equation}
where $\lambda = n(1-s) / \bar{P}$ linearly scales the raw profile values so that the unclamped mean fan-in matches the target $(1-s)n$. After clamping to $[f_{\min},\, n]$, the realized total may exceed the target and we report both target and realized sparsity. For lognormal and powerlaw profiles, where clamping substantially distorts heavy tails, $\lambda$ is instead found by binary search to achieve the target mean fan-in after clamping. We use $f_{\min} = 1$ in all experiments. The minimum fan-in prevents complete disconnection at extreme sparsity. Because $f_{\min}$ increases the total parameter count, we distinguish target sparsity $s_{\text{target}}$ from realized sparsity $s_{\text{actual}}$ and report both.

The minimum fan-in imposes a ceiling on achievable sparsity:
\begin{equation}
S_{\max} = 1 - \frac{f_{\min}}{M_{\text{in}}}.
\end{equation}
For the 784-input datasets, $S_{\max} = 99.87\%$ at $f_{\min}=1$, comfortably above all test points. For Forest Cover ($M_{\text{in}}=54$), $S_{\max} = 98.15\%$ at $f_{\min}=1$; we therefore limit Forest Cover experiments to 80--98\% sparsity (Table~\ref{tab:sparsity-max}).

\begin{table}[h!]
    \centering
    \caption{Maximum achievable sparsity ($S_{\max}$) as a function of minimum fan-in and input dimension.}
    \label{tab:sparsity-max}
    \small
    \renewcommand{\arraystretch}{1.2}
    \begin{tabularx}{0.80\textwidth}{l >{\centering\arraybackslash}X >{\centering\arraybackslash}X}
        \toprule
        \textbf{Min fan-in} & \textbf{784-input datasets} & \textbf{Forest Cover} ($M_{\text{in}}\!=\!54$) \\
        \midrule
        1  & 99.87\% & 98.15\% \\
        5  & 99.36\% & 90.74\% \\
        10 & 98.72\% & 81.48\% \\
        20 & 97.45\% & 62.96\% \\
        \bottomrule
    \end{tabularx}
\end{table}

At extreme sparsity the $f_{\min}$ constraint compresses the fan-in distribution. For the exponential profile on MNIST FC1, the realized fan-in ratio is approximately 20$\times$ at 90\% sparsity (247 to 12 connections). At 99.9\% sparsity it collapses to just 1.2$\times$ as nearly all neurons are clamped to $f_{\min}=1$.

\subsection{Input Spreading and Mask Construction}

Given fan-in values $\{f_i\}$, we must select which input neurons connect to each output neuron. Even spreading deterministically distributes connections:
\begin{equation}
j_k = \left\lfloor \left(\phi_i + \frac{k \, n}{f_i}\right) \bmod n \right\rfloor, \quad k = 0, \dots, f_i - 1,
\end{equation}
where $\phi_i = \lfloor i \cdot \varphi \cdot n \rfloor \bmod n$ is a per-neuron offset based on the golden ratio $\varphi \approx 1.618$. Without the offset, all neurons with the same fan-in would select the same subset of inputs, creating highly non-uniform fan-out. The golden-ratio offset ensures that different neurons sample maximally dispersed subsets, yielding approximately uniform fan-out across inputs. Random spreading samples $f_i$ inputs uniformly without replacement. Sequential spreading (inputs $0, \ldots, f_i-1$) causes catastrophic input coverage bias (75.0\% vs.\ 97.5\% test accuracy at 90\% sparsity on MNIST) and is not used in experiments.

Connectivity is implemented via binary masks $M \in \{0,1\}^{m \times n}$ applied to dense weights: $W_{\text{sparse}} = M \odot W$. For static PSN experiments, masks are fixed at initialization and reapplied after each gradient update. For RigL experiments, masks are periodically updated.

\subsection{Initialization in Heterogeneous Sparse Networks}
Standard initialization schemes (Xavier\citep{Glorot2010}, He\citep{He2015}) assume homogeneous fan-in across neurons. In PSN, fan-in varies deterministically across neurons, creating a tension between two initialization objectives.

For a neuron $i$ with fan-in $f_i$ and inputs with variance $\sigma_x^2$, the pre-activation variance is $\mathrm{Var}(z_i) = f_i \sigma_w^2 \sigma_x^2$. A per-neuron scaling $\sigma_w^2 = 1/f_i$ preserves forward activation variance across neurons. However, the gradient with respect to an input $x_j$ is
\[
\frac{\partial L}{\partial x_j} = \sum_{i : j \in \mathcal{I}(i)} W_{ij} \frac{\partial L}{\partial z_i},
\]
whose variance scales with the fan-out of $x_j$. Under even spreading, fan-out is approximately uniform ($\approx m\mathbb{E}[f]/n$), suggesting a global scaling $\sigma_w^2 \propto 1/\mathbb{E}[f]$ instead to stabilize backward propagation.

These two conditions are incompatible when fan-in is heterogeneous. The per-neuron scaling preserves forward statistics but induces heterogeneous gradient amplification at hub neurons with large $f_i$, while mean fan-in scaling equalizes backward statistics but does not preserve per-neuron forward variance. In initial experiments, per-neuron scaling $\sigma_w^2 = 2/f_i$ produced unstable optimization with large gradient norms and occasional divergence, particularly at high sparsity. Mean fan-in scaling produced stable training dynamics across all seeds and sparsity levels tested. We therefore adopt He initialization\citep{He2015} with mean fan-in:
\begin{equation}
\sigma_w^2 = \frac{2}{\mathbb{E}[f]} = \frac{2}{(1 - s_{\text{actual}})\, n},
\end{equation}
where the factor of 2 accounts for the ReLU activation zeroing approximately half of the pre-activations. In practice, LayerNorm re-normalizes activations before each ReLU, so the distinction between He ($2/\mathbb{E}[f]$) and Xavier ($1/\mathbb{E}[f]$) initialization has minimal impact on training dynamics. The critical choice is using mean fan-in rather than per-neuron fan-in. Mean fan-in initialization and LayerNorm set stable gradient magnitudes at the start of training, and maintain consistent activation scaling throughout, respectively. Without correct initialization, early gradient steps can be destabilising even with LayerNorm. Without LayerNorm, fan-in heterogeneity causes systematic activation imbalance that initialization alone cannot prevent. This should be understood as a practical heuristic rather than a theoretically derived optimum. It prioritizes stable gradient propagation at the cost of non-uniform forward variance across neurons. Whether a more nuanced initialization (e.g., fan-in-dependent scaling with gradient clipping) could improve performance at extreme sparsity remains unverified.

\subsection{Activation Functions and Normalization}\label{sec:layernorm}

All hidden layers apply Layer Normalization\citep{Ba2016} followed by a Rectified Linear Unit activation ($\mathrm{LN} \to \mathrm{ReLU}(z) = \max(0, z)$). LayerNorm normalizes the pre-activation vector to zero mean and unit variance across features, decoupling activation scale from fan-in. This is particularly important in heterogeneous sparse networks, where hub neurons with large fan-in would otherwise produce systematically larger pre-activations than specialists with small fan-in. Without LayerNorm, high-sparsity runs on Fashion-MNIST and Forest Cover exhibited training instability and high seed variance. Empirically, LayerNorm equalises activation magnitudes across neurons regardless of fan-in and the hub and specialist neurons show approximately equal mean activation magnitudes (${\sim}1\times$ ratio), confirming that low-fan-in specialists are not rendered inactive by their sparse connectivity. Bias terms are initialized to zero.

\subsection{Training Procedure}\label{sec:training}

All networks are trained with the Adam optimizer ($\eta = 0.001$, default $\beta$), cross-entropy loss, and evaluated on held-out test sets without a validation split. Static experiments train for 10 epochs on MNIST, 10 on Fashion-MNIST, 20 on EMNIST-Balanced, and 30 on Forest Cover, RigL experiments use 20 epochs on the 784-input datasets and 30 on Forest Cover. Hyperparameters are fixed across all conditions within each dataset. Sparse masks are generated at initialization according to the specified profile, spreading pattern, and sparsity level. Masks remain fixed throughout training and are reapplied after each gradient update. This enables deterministic reproducibility and clean isolation of topological effects. We test the parametric profiles and a random-uniform baseline, each with both even and random spreading, across multiple sparsity levels. Each configuration uses 5 random seeds (42, 123, 456, 789, 1024).

We also evaluate PSN-style lognormal connectivity as initialisation for RigL,\citep{Evci2020} a dynamic sparse training method that periodically updates the mask by removing low-magnitude weights and growing connections along large gradient directions. RigL hyperparameters used were drop fraction $\alpha = 0.3$, update period $\Delta T = 100$ steps, cosine drop-fraction schedule terminating at $T_{\mathrm{end}} = 0.75 \times T_{\mathrm{total}}$ steps. The output layer is excluded from mask updates. We compare ERK (Erd\H{o}s-R\'enyi-Kernel), uniform, and lognormal initialization strategies at a specified target CCV.  RigL converges to a characteristic fan-in CCV that depends on the network architecture and sparsity, but not on the task or initialisation (Table~\ref{tab:equil-ccv}). The 784-input datasets (MNIST, Fashion-MNIST, EMNIST) share the same architecture and therefore cluster at nearly identical equilibrium CCVs. Forest Cover's small FC1 input dimension (54) reduces the achievable CCV by 0.4--0.6 units. We use the equilibrium CCV measured at each (dataset, sparsity) combination to set the lognormal target CCV for initialisation.

\begin{table}[h!]
    \centering
    \caption{Equilibrium fan-in CCV reached by RigL (default hyperparameters, 5 seeds). MNIST, Fashion-MNIST, and EMNIST share the same 784$\to$1024$\to$1024 architecture and converge to nearly identical CCVs. Forest Cover is lower at all sparsities due to the 54-input FC1 bottleneck.}
    \label{tab:equil-ccv}
    \small
    \renewcommand{\arraystretch}{1.2}
    \begin{tabularx}{0.90\textwidth}{l >{\centering\arraybackslash}X >{\centering\arraybackslash}X >{\centering\arraybackslash}X >{\centering\arraybackslash}X}
        \toprule
        \textbf{Dataset} & \textbf{80\%} & \textbf{90\%} & \textbf{95\%} & \textbf{98\%} \\
        \midrule
        MNIST            & 1.62 & 2.43 & 3.48 & 5.49 \\
        Fashion-MNIST    & 1.68 & 2.52 & 3.61 & 5.71 \\
        EMNIST-Balanced  & 1.68 & 2.53 & 3.64 & 5.69 \\
        Forest Cover     & 1.26 & 2.10 & 3.17 & 5.13 \\
        \bottomrule
    \end{tabularx}
\end{table}

\subsection{Multi-Peak Profile Interpolation}

To test whether gradient hierarchy arises from fan-in heterogeneity in general (rather than specific profile shapes), we define a family of multi-peak profiles that continuously interpolates between maximal heterogeneity and uniformity. The density at normalized index $t \in [0,1]$ is:
\begin{equation}
    P_k(t) = \max_{j=0,\ldots,k-1} \exp\!\left(-\beta \left(t - c_j\right)^2\right), \quad c_j = \frac{2j+1}{2k},
\end{equation}
where $k$ evenly-spaced Gaussian peaks of sharpness $\beta$ cover the neuron index axis. At $k=1$, a single peak creates one dense hub cluster. As $k \to \infty$, the peaks merge into a uniform density, recovering the random-uniform baseline. The fan-in CCV decreases monotonically with $k$ for fixed $\beta$. We vary $k \in \{1, 2, \ldots, 20, 25, 30, 40, 50\}$ and $\beta \in \{12, 50, 1200\}$, all at 90\% sparsity with even spreading and 5 seeds (360 total runs).

\subsection{Evaluation Metrics}

The primary metric is test accuracy, reported as mean $\pm$ standard deviation over five seeds (42, 123, 456, 789, 1024). Realized sparsity (fraction of zero weights in hidden layers) is reported alongside target sparsity to account for minimum fan-in constraints. For RigL experiments, we additionally report the fan-in CCV at convergence (measured over the final mask state averaged across seeds).

To quantify gradient concentration at high-connectivity neurons, we compute the gradient ratio, defined as the mean absolute gradient magnitude of the top-half neurons by fan-in divided by that of the bottom half. Specifically, neurons in each layer are ranked by fan-in and split at the median into hub (above median) and specialist (below median) groups. The ratio is then $\bar{|\nabla W|}_{\mathrm{hub}} / \bar{|\nabla W|}_{\mathrm{spec}}$, measured after one forward-backward pass on a training batch. This rank-based split is an arbitrary summary statistic of a continuous relationship between fan-in and gradient magnitude. There are no distinct populations, and the 50\% threshold has no privileged status. We adopt it as a convenient scalar measure of gradient heterogeneity that correlates strongly with fan-in CV ($r = 0.93$). Importantly, gradient concentration at hub neurons is a structural scaling property of the mask geometry, not a functional advantage and it does not translate into accuracy differences on the tasks tested here.

\section{Results and discussion}

\subsection{MNIST dataset}

MNIST serves as a diagnostic tool for probing the internal dynamics induced by structured sparse connectivity. The two-hidden-layer fully-connected network with 1,024 neurons per hidden layer (1,867,786 total parameters) achieved 97.81\% $\pm$ 0.18\% test accuracy on MNIST digit classification after training for 5 epochs with Adam optimizer at learning rate 0.001 and batch size 128. All five seeds (42, 123, 456, 789, 1024) converged to $>$95\% test accuracy within the first epoch, with individual final accuracies ranging from 97.66\% to 98.12\%. The tight standard deviation of 0.18\% across seeds confirms high reproducibility. The 97.81\% figure establishes the baseline for sparse variants to approach while using substantially fewer active parameters. Mean fan-in initialization enables stable training across all sparsity levels tested (up to 99.9\%). We did not systematically compare per-neuron initialization in this study (the mean fan-in approach was adopted early based on initial experiments in which per-neuron scaling led to training instability at high sparsity). All models are trained for 5 epochs and the convergence is rapid, with most runs at sparsity levels up to 99.9\% converging within the first epoch. Overfitting is negligible and the mean gap between best and final test accuracy is 0.04\% across all runs, with a maximum of 0.71\%. The rapid convergence reflects that MNIST is a relatively easy task for networks of this scale, a characteristic that shapes the interpretation of sparsity experiments and the hub--specialist gradient hierarchy which emerges within a single epoch and persists throughout training.

We first compared eight connectivity profile variants at different sparsities using both even and random input spreading, with minimum fan-in constraint of one. The profiles tested were quadratic, exponential, linear, and bell (standard orientations), plus their inverse counterparts (inverse quadratic, inverse exponential, inverse linear, inverse bell), where hub and specialist roles are swapped. Each configuration was tested with both even and random spread patterns, and trained with five independent random seeds, yielding 80 PSN configurations at each sparsity level plus a random-uniform baseline. Figure~\ref{fig:fan-in-profiles}c illustrates a representative PSN mask for the exponential profile with random spreading at 90\% sparsity (FC1 layer). The exponential decay assigns fan-in ranging from 247 (densely connected neurons at low indices) to 12 (sparsely connected neurons at high indices), a 20.6$\times$ ratio. The mean fan-in is 77.9, consistent with the target 10\% density (78.4 = 784 $\times$ 0.1). Random spreading distributes connections stochastically across the input dimension. All 784 input pixels receive between 74 and 125 connections (std = 8.7), with zero dead inputs.

During initial experiments, we implemented input spreading using sequential assignment where output neuron $i$ connected to input neurons $[0, 1, \ldots, f_i - 1]$ in contiguous blocks. This created catastrophic input coverage bias that collapsed performance by over 22\%. In a layer with exponential profile at 90\% sparsity, early output neurons have high fan-in ($\sim$100 connections) while late output neurons have fan-in approaching 1. With sequential spreading, output neuron 0 connects to inputs $[0, 99]$, output neuron 1 connects to inputs $[0, 98]$, and so on. Summing across all output neurons, input neuron 0 receives connections from nearly all 1,024 output neurons, while late input neurons receive connections only from the few output neurons with the highest fan-in. This creates a 1,024$\times$ fan-out ratio. The most-connected input neuron has 1,024 outgoing connections versus the least-connected having near zero. Worse, 537 of 784 input neurons (69\%) receive zero connections from any output neuron, they are completely dead inputs that the network cannot learn from. We compared four conditions at 90\% sparsity with exponential profile on MNIST (10 epochs, seed 42) and found that sequential spreading degrades accuracy by 22.49\% compared to even spreading, a catastrophic failure caused entirely by the input coverage bias. The network can only learn from the first $\sim$250 input pixels while ignoring the remaining 534, regardless of which pixels contain discriminative features for digit recognition. Even spreading distributes connections uniformly ensuring every input neuron receives approximately equal total connectivity. Random spreading achieves similar balance stochastically. Both yield performance within 0.3\% of each other confirming that the poor performance is specific to sequential assignment.  

The sparsity sweep reveals remarkable resilience of all architectures to sparsity. All models, including the unstructured random baseline, achieve nearly identical performance. Test accuracy remains above 97\% for all models up to 98\% sparsity, and above 96\% at 99\% sparsity. The spread across all model variants (8 profiles × 2 patterns + random) spans only around 0.2\%, with fully overlapping confidence intervals. No pairwise comparison reaches statistical significance (all p > 0.1). The comparison between standard and inverse profile orientations shows no systematic difference (data not shown) across all sparsities (maximum absolute difference of 0.1--0.2\%). Inverting the fan-in assignment changes which neurons receive many connections versus few, but does not affect task performance. 

We tested also whether the spatial arrangement of hub neurons, evenly spaced throughout a layer versus randomly scattered, affects model performance and found negligible differences with largest gap of 0.34\% with the mean absolute difference of just 0.11\%. In the following we will show the results for random spreading of inputs.

\begin{table}[h!]
    \centering
    \caption{Test accuracy (\%) for even-spread PSN profiles and random baseline across sparsity levels. Values are means over 5 seeds. Dense baseline: 97.81\% $\pm$ 0.18\%.}
    \label{tab:sparsity-results}
    \small
    \renewcommand{\arraystretch}{1.2}
    \begin{tabularx}{\textwidth}{l *{5}{>{\centering\arraybackslash}X}}
        \toprule
        \textbf{Sparsity} & \textbf{Quadratic} & \textbf{Exponential} & \textbf{Linear} & \textbf{Bell} & \textbf{Random} \\
        \midrule
        85\%   & 97.77 & 97.69 & 97.79 & 97.76 & 97.87 \\
        88\%   & 97.88 & 97.67 & 97.73 & 97.78 & 97.86 \\
        90\%   & 97.79 & 97.78 & 97.77 & 97.76 & 97.84 \\
        92\%   & 97.75 & 97.76 & 97.76 & 97.69 & 97.78 \\
        94\%   & 97.61 & 97.55 & 97.60 & 97.55 & 97.74 \\
        96\%   & 97.47 & 97.47 & 97.48 & 97.42 & 97.38 \\
        98\%   & 97.21 & 97.09 & 97.21 & 97.11 & 97.19 \\
        99\%   & 96.76 & 96.52 & 96.66 & 96.61 & 96.93 \\
        99.9\% & 92.49 & 92.76 & 92.50 & 92.54 & 92.49 \\
        \bottomrule
    \end{tabularx}
\end{table}

Degradation is gradual and uniform. From 85\% to 98\% sparsity, accuracy drops by only $\sim$0.6\% across all models. The steep decline occurs only in the final step from 99\% to 99.9\%, where accuracy drops approximately 4\%. This cliff-like behavior at 99.9\% suggests a critical threshold where network capacity becomes genuinely limiting. At 90\% sparsity, hidden layers retain approximately 80,800 (FC1) and 105,400 (FC2) active connections out of $\sim$803,000 and $\sim$1,049,000 total, respectively. This represents substantial connectivity and each neuron receives on average $\sim$79 (FC1) or $\sim$103 (FC2) inputs. Even at 99.9\% sparsity, where the $f_{\min}=1$ constraint forces most neurons to receive exactly one input connection ($\sim$1,024 active connections per layer), all models still achieve approximately 92.5\% accuracy, far exceeding chance level. At this point the intended profile heterogeneity collapses and all neurons become functionally identical, explaining why all profiles converge to the same accuracy.

\begin{figure}[h!]
    \centering
    \includegraphics[width=0.75\textwidth]{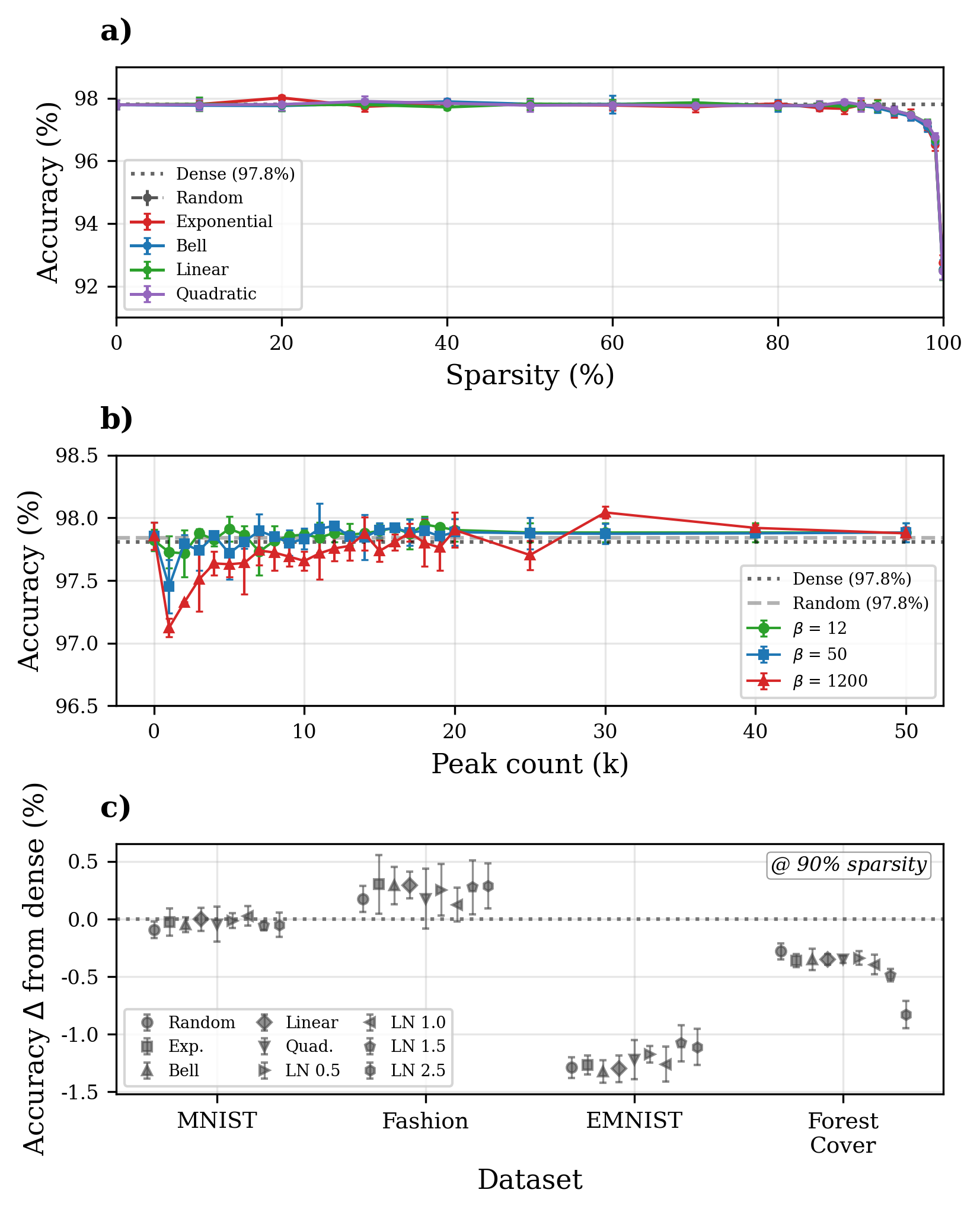}
    \caption{Static connectivity structure does not affect accuracy. \textbf{(a)} MNIST accuracy versus sparsity for four PSN profiles and the random baseline (5 seeds each). All profiles cluster within 0.2\% of each other up to 98\% sparsity. The steep decline at 99.9\% reflects capacity exhaustion (mean fan-in collapses to $\sim$1), not profile-specific effects. \textbf{(b)} Multi-peak interpolation at 90\% sparsity on MNIST. Peak count $k$ varies fan-in CCV continuously from 3.2 ($k=1$) to 0 ($k=50$) for three sharpness values $\beta$. Accuracy is flat across the entire range (total spread 0.65\%). The only visible deviation occurs at $\beta = 1200$, $k = 1$--$2$, where extreme concentration of fan-in into $\sim$100 densely connected neurons produces a 0.7\% drop.\textbf{(c)} Accuracy relative to dense baseline at 90\% sparsity across all four datasets. Each marker represents one profile (9 profiles per dataset, 5 seeds). All profiles fall within $\pm$0.5\% of the dense baseline except lognormal CV\,=\,2.5 on Forest Cover, where the 54-dimensional input limits useful heterogeneity.}
    \label{fig:null-result}
\end{figure}

At moderate sparsity (90\%), the $f_{\min}=1$ floor is rarely binding and the mean fan-in across profiles is 78 (FC1) to 102 (FC2), with minimums averaging 4.5 to 5.9.

Inverse profiles confirm that spatial arrangement does not affect task performance. For each profile, its inverse (swapping which neurons receive high versus low fan-in) achieves identical accuracy (mean difference: 0.01\%, $p = 0.57$). This indicates that the statistical distribution of fan-in values, not the spatial position of high-connectivity neurons within the layer, determines outcomes. Profile orientation can therefore be chosen freely without performance consequences.

Each profile has a characteristic fan-in coefficient of variation (CV = $\sigma_f / \mu_f$). Exponential has CV $\approx 0.82$, bell $\approx 0.66$, linear $\approx 0.45$, quadratic $\approx 0.25$, and the random baseline $\approx 0$. These values remain stable across sparsity levels 85\%--98\% for each profile, collapsing together at 99.9\% when the minimum fan-in constraint forces all neurons to near-identical connectivity.

To test whether accuracy depends on the degree of fan-in heterogeneity rather than any specific profile shape, the multi-peak family continuously varies fan-in CV from $\approx 3.2$ ($k=1$, $\beta=1200$, a single narrow hub cluster) to $\approx 0$ ($k=50$, approximately uniform). Fan-in CV decreases monotonically with $k$ for each $\beta$, with the transition rate determined by how quickly peak overlap erases the valleys in the fan-in distribution.

Figure~\ref{fig:null-result}b shows accuracy as a function of $k$ for all three $\beta$ values. Accuracy is essentially flat across the entire parameter space. The total accuracy range across all 72 multi-peak configurations (24 $k$-values $\times$ 3 $\beta$-values) spans only 0.65\% (97.12\%--97.78\%), compared to the dense baseline of 97.81\% $\pm$ 0.18\% and random baseline of 97.84\% $\pm$ 0.14\%. The accuracy spread within each $\beta$ value is even smaller: 0.25\% for $\beta = 12$, 0.21\% for $\beta = 50$, and 0.63\% for $\beta = 1200$. The only measurable accuracy effect occurs at $\beta = 1200$, $k = 1$, where the extreme concentration of connectivity (fan-in CV $= 3.24$) produces a 0.69\% accuracy drop. Even this configuration, where $\sim$900 of 1024 neurons receive a single input connection and a handful receive all 784, still achieves 97.12\% $\pm$ 0.18\%. This is consistent with the conclusion that MNIST is saturated with respect to connectivity structure: the task is too easy for the capacity of a 1024-unit hidden layer, regardless of how that capacity is distributed.

\subsection{Cross-dataset static evaluation}

To enable direct cross-dataset comparison, we trained all four datasets with matched settings (10 epochs for the 784-input datasets, 30 for Forest Cover, 5 seeds per configuration). Dense baselines: MNIST 98.16\% $\pm$ 0.12\%, Fashion-MNIST 89.03\% $\pm$ 0.27\%, EMNIST-Balanced 85.97\% $\pm$ 0.15\%, Forest Cover 95.60\% $\pm$ 0.09\%. Note that the MNIST baseline here (98.16\%, 10 epochs) is slightly higher than the 97.81\% reported in the diagnostic section above (5 epochs); results within each comparison use a consistent training protocol.

\begin{table}[h!]
    \centering
    \caption{Test accuracy (\%) at 90\% sparsity for static PSN profiles across four datasets. Values are means over 5 seeds. Profiles are listed in order of increasing fan-in CV. ``Random'' denotes uniform random connectivity (CV $\approx$ 0).}
    \label{tab:cross-dataset-static}
    \small
    \renewcommand{\arraystretch}{1.2}
    \begin{tabular}{l c c c c}
        \toprule
        \textbf{Profile (CV)} & \textbf{MNIST} & \textbf{Fashion} & \textbf{EMNIST} & \textbf{Forest} \\
        \midrule
        Dense & 98.16 & 89.03 & 85.97 & 95.60 \\
        \midrule
        Random ($\approx$\,0) & 98.07 & 89.20 & 84.68 & 95.32 \\
        Quadratic ($\approx$\,0.25) & 98.12 & 89.20 & 84.74 & 95.25 \\
        Bell ($\approx$\,0.66) & 98.11 & 89.32 & 84.64 & 95.25 \\
        Exponential ($\approx$\,0.82) & 98.14 & 89.33 & 84.70 & 95.24 \\
        Lognormal (1.0) & 98.19 & 89.15 & 84.71 & 95.21 \\
        Lognormal (1.5) & 98.10 & 89.30 & 84.89 & 95.11 \\
        Lognormal (2.5) & 98.11 & 89.31 & 84.86 & 94.77 \\
        \bottomrule
    \end{tabular}
\end{table}

The MNIST result replicates across all datasets. No static profile achieves a statistically significant improvement over the random baseline at $p < 0.05$. The total accuracy spread across profiles is 0.20\% (MNIST), 0.18\% (Fashion-MNIST), 0.25\% (EMNIST), and 0.55\% (Forest Cover), all within inter-seed variability. On Forest Cover, higher-CV profiles show a small deficit. Lognormal CV\,=\,2.5 achieves 94.77\% versus random at 95.32\%, likely because the 54-dimensional input layer limits the useful range of fan-in heterogeneity. At 90\% sparsity, the mean fan-in in the first hidden layer is only 5.4, leaving little room for meaningful differentiation between high- and low-connectivity neurons. At 98\% sparsity on Fashion-MNIST (3 seeds, separate experiment), the same pattern holds with powerlaw CV\,=\,1.0 achieving the best accuracy (88.56\%) versus random (88.16\%), a +0.40\% gap that does not reach statistical significance.

\subsection{Dynamic sparse training with PSN initialization}

The static experiments establish that connectivity structure alone does not improve accuracy on the tested tasks. We next asked whether PSN fan-in profiles might serve as better starting points for dynamic sparse training. Using RigL, we initialized masks with lognormal fan-in distributions at the measured equilibrium CCV and $\pm$0.5 offsets, and compared against ERK and uniform initialization across sparsity levels from 80\% to 98\%.

\begin{figure}[h!]
    \centering
    \includegraphics[width=\textwidth]{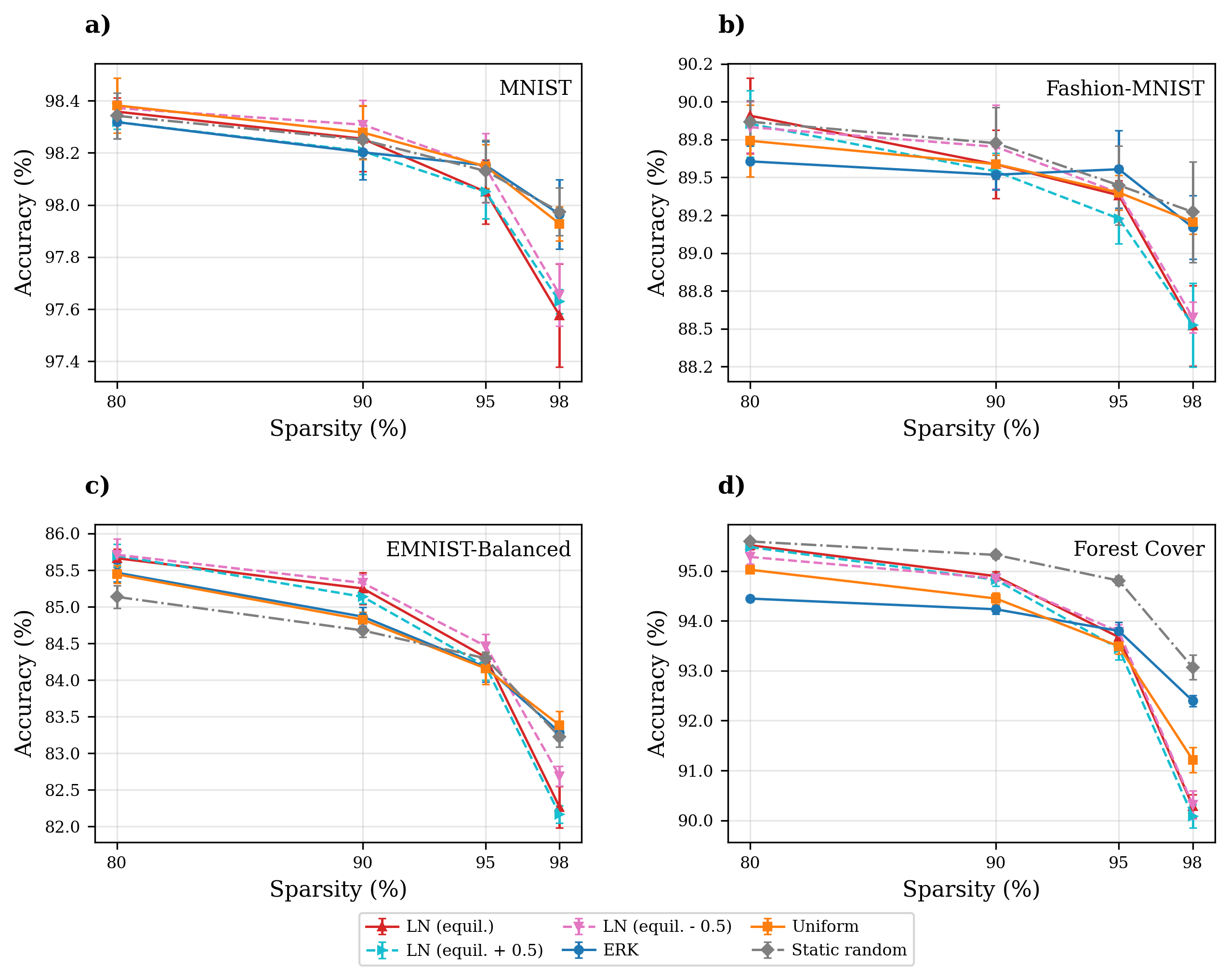}
    \caption{RigL accuracy versus sparsity by initialisation strategy across four datasets (5 seeds each). Six initialisation strategies are compared at each sparsity level. LN (equil.) denotes lognormal initialisation matched to the measured equilibrium fan-in CCV for each (dataset, sparsity) combination. LN (equil. $\pm$ 0.5) offsets the target CCV by 0.5 above or below equilibrium. \textbf{(a)} On MNIST, all strategies perform within 0.5\% of each other. \textbf{(b)} On Fashion-MNIST, LN (equil.) maintains a consistent edge over ERK and uniform across all sparsity levels. \textbf{(c)} On EMNIST-Balanced, separation between strategies increases at higher sparsity. \textbf{(d)} On Forest Cover, ERK collapses at 95--98\% sparsity due to the 54-input bottleneck in FC1. Static random (dashed gray) outperforms ERK at all sparsity levels on this dataset.}
    \label{fig:rigl-accuracy}
\end{figure}

Figure~\ref{fig:rigl-accuracy} shows accuracy versus sparsity for all six initialisation strategies across the four datasets. A consistent ordering emerges. Lognormal initialisation at the equilibrium CCV achieves the highest or near-highest accuracy across all datasets and sparsity levels, followed by ERK and uniform, with the gap widening at higher sparsity. On harder tasks, the advantage of equilibrium-matched initialisation grows. At 90\% sparsity the gap between lognormal (equil.) and ERK is +0.16\% on Fashion-MNIST, +0.43\% on EMNIST, and +0.49\% on Forest Cover.

The Fashion-MNIST result was confirmed with 10 independent seeds ($p = 0.036$, Cohen's $d = 1.07$ versus ERK). The ordering across initialization strategies is monotonic with proximity to the equilibrium CCV. At 90\% sparsity, RigL converges to CCV $\approx$ 2.5 regardless of initialization (Table~\ref{tab:equil-ccv}), and starting at this equilibrium point produces the best final accuracy. Examination of training curves reveals that lognormal CV\,=\,2.5 starts below other initializations in early epochs, then crosses over at approximately epoch 10--12 to finish highest. This is consistent with the topology already being near equilibrium, allowing the optimizer to focus on refining weight values rather than rearranging connections.

On Forest Cover, ERK (94.27\%) substantially underperforms all other RigL configurations, including static random (94.75\%). ERK allocates sparsity proportional to layer dimensions, which assigns the 54-input first hidden layer a mean fan-in of only $\sim$5, effectively starving the input layer. Lognormal initialization distributes heterogeneity within each layer independently and avoids this bottleneck.

\section{Review of related work}

The sparse neural network literature has extensively explored weight pruning,\citep{LeCun1990, Han2015} dynamic evolution of topology,\citep{Mocanu2018, Evci2020} and structured approaches including block sparsity\citep{Gray2017} and mixture-of-experts architectures.\citep{Shazeer2017} Early work demonstrated that trained neural networks contain substantial redundancy. Optimal Brain Damage\citep{LeCun1990} and Optimal Brain Surgeon\citep{Hassibi1993} used second-order derivative information to identify and remove low-saliency weights with minimal impact on performance, showing that principled weight selection outperforms naive magnitude-based deletion. Han et al. later showed that simple iterative magnitude pruning can achieve 9--13$\times$ compression on AlexNet and VGG while maintaining ImageNet accuracy.\citep{Han2015} A key limitation of magnitude pruning is the requirement for expensive dense pretraining before sparsification. The Lottery Ticket Hypothesis revealed that dense networks contain sparse subnetworks (lottery tickets) that, when trained from their original initialization, match full network performance.\citep{Frankle2019} This finding suggests successful sparse training depends on fortunate initialization rather than requiring dense pretraining, though identifying the winning tickets requires expensive pruning-retraining cycles. One-shot pruning methods eliminate the iterative search. Single-shot Network Pruning (SNIP)\citep{lee2018snip} uses connection sensitivity computed on a single batch to prune before training, but suffers from layer collapse at high sparsity where entire layers can be completely pruned. Iterative Synaptic Flow Pruning (SynFlow) addresses this through iterative data-free pruning based on gradient flow, requiring 100+ forward passes but avoiding layer collapse through progressive refinement.\citep{tanaka2020pruning} 

Rather than finding fixed sparse topologies, dynamic methods evolve connectivity during training. Sparse Evolutionary Training (SET) removes smallest-magnitude weights and regrows connections randomly, discovering that networks naturally evolve toward scale-free topologies with heterogeneous degree distributions despite uniform random initialization.\citep{Mocanu2018} Rigging the Lottery (RigL) refined dynamic sparse training by using gradient magnitudes to guide growth rather than random selection.\citep{Evci2020} During each topology update, RigL removes the smallest magnitude weights and regrows weights at positions with highest gradient magnitudes. This approach consistently outperforms random growth strategies across diverse tasks and network architectures. Lasby et al.\citep{lasby2024dynamic} further showed that RigL evolves high fan-in variance and naturally ablates entire neurons at high sparsity, and that imposing constant fan-in constraints harms performance unless neuron ablation is explicitly re-enabled.

A common thread across the existing methods is their reliance on uniform or uniform random sparsity patterns. Magnitude pruning removes the smallest weights globally or per-layer without considering neuron-level organization. Random initialization in SET and gradient-guided growth in RigL operate on individual weights without explicit constraints on per-neuron connectivity distributions. The Lottery Ticket Hypothesis searches for winning subnetworks within randomly initialized dense networks, inheriting the statistical uniformity of random initialization. While these approaches achieve impressive compression results, they do not explore whether designing structured heterogeneity from initialization provides advantages over discovering or inheriting random structure. 

Rather than unstructured weight-level sparsity, structured approaches organize zeros in patterns aligned with hardware or architectural constraints. Block-sparse networks group weights into blocks for efficient computation on modern accelerators.\citep{Gray2017} Convolutional networks naturally employ structured sparsity through local receptive fields and weight sharing,\citep{LeCun1998} a property that can be exploited through low-rank approximations.\citep{Denton2014} MobileNets employ depthwise separable convolutions that factorize standard convolutions into depthwise and pointwise operations, reducing parameters while maintaining expressive power.\citep{Howard2017} EfficientNets scale network width, depth, and resolution jointly to optimize efficiency-accuracy tradeoffs.\citep{Tan2019}

Mixture-of-Experts architectures create dynamic sparsity by routing inputs to subsets of expert networks, enabling massive scale while maintaining per-example computational costs.\citep{Shazeer2017}  During training and inference, only a small fraction of the total parameters (typically 2-4 experts out of hundreds or thousands) processes each example, creating input-dependent sparsity patterns. Recent Switch Transformers scaled this to trillion-parameter models.\citep{Fedus2022} However, these methods operate at module or block/module level rather than introducing fine-grained per-neuron heterogeneity within individual fully-connected layers.

Group-structured sparsity methods assign weights to groups and apply shared sparsity patterns within each group. Grouped convolutions partition input and output channels into independent groups that process data separately before concatenation, reducing parameters proportionally to group count. Neural architecture search methods discover efficient network structures including layer widths, depths, and connectivity patterns through reinforcement learning.\citep{Zoph2017} While these approaches demonstrate the importance of network structure, they optimize at the architectural design level rather than exploring systematic organization of heterogeneous connectivity within layers.

Recent work on attention mechanisms introduces sparse connectivity patterns that emerge from learned attention weights. Sparse attention in Transformers restricts attention computation to subsets of positions using fixed patterns such as local windows or strided patterns, reducing the quadratic computational complexity of standard attention.\citep{Child2019} Routing transformers learn to cluster tokens and attend only within clusters, creating dynamic sparse attention patterns.\citep{Roy2021} These methods demonstrate benefits of structured rather than uniform sparsity in attention layers, though they address the specific computational characteristics of self-attention rather than general feed-forward layers.

The Cannistraci group researched brain-inspired sparse topology design. Zhang et al. introduced Cannistraci-Hebb Training (CHT), which combines a brain-inspired Bipartite Receptive Field (BRF) initialization with gradient-free topology-driven link regrowth.\citep{zhang2026brain} At 99\% sparsity, CHT outperforms other DST methods and fully connected networks on MLP image classification tasks. Cerretti et al. extended this with the Dendritic Network Model (DNM), which generates sparse topologies through parametric distributions of dendrites, receptive fields, and degree, enabling control over modularity and degree heterogeneity.\citep{Cerretti2025dendritic} DNM outperforms classical sparse initializations at 99\% sparsity across MNIST, Fashion-MNIST, EMNIST, and CIFAR-10 in both static and dynamic sparse training.

Several other recent methods address sparse initialization quality without explicit heterogeneity control. Exact Orthogonal Initialization (EOI) constructs sparse weight matrices that are exactly orthogonal at initialization, preserving signal norms through sparse layers.\citep{Nowak2024eoi} EOI demonstrates that initialization quality matters enormously for static sparse training, achieving substantial improvements over standard Kaiming initialization at high sparsity. Ma et al. recently demonstrated that static sparse connectivity can be effective in deep reinforcement learning.\citep{Ma2025}

PSN occupies a specific position in this emerging landscape. Like DNM, it designs heterogeneous connectivity from initialization. Unlike DNM, PSN parameterizes connectivity through continuous profile functions evaluated at neuron indices, yielding deterministic, reproducible, and continuously tunable fan-in distributions. This enables controlled experiments that isolate fan-in heterogeneity (measured by CV) as a single scalar variable. Continuous control can be achieved by smoothly interpolating from maximal heterogeneity to uniformity within a single parameterized family of profiles.

\begin{sidewaystable*}[htbp]
\begin{threeparttable}
    \centering
    \scriptsize 
    \caption{Comprehensive Comparison of Sparse Training Methods}
    \label{tab:sparse_comparison}
    \setlength{\tabcolsep}{3pt} 
    \renewcommand{\arraystretch}{1.8} 
    
    \begin{tabularx}{\textwidth}{@{} l l l >{\raggedright\arraybackslash}X l c >{\raggedright\arraybackslash}X >{\raggedright\arraybackslash}X >{\raggedright\arraybackslash}X l @{}}
        \toprule
        \textbf{Method} & \textbf{Net Type} & \textbf{Dynamics} & \textbf{Topology} & \textbf{Sparsification Method} & \textbf{Data} & \textbf{Overhead} & \textbf{Heterogeneity} & \textbf{Structural Origin} & \textbf{Purpose} \\
        \midrule
        
        \textbf{Weight Pruning} \citep{LeCun1990, Han2015} & Feedforward & Static & Dense to Sparse Uniform & Magnitude / Saliency & Full & Low & Emergent & Post-training & Compression \\
        
        \textbf{Lottery Ticket} \citep{Frankle2019} & Any & Static & Dense to Sparse Uniform & Magnitude + Rewind & Full & Very High & Inherited & Initial Luck & Find Subnetworks \\
              
        \textbf{SET} \citep{Mocanu2018} & Feedforward & Dynamic & Random to Scale-free & Magnitude & Full & High (rewiring) & Emergent & Evolution & Evolve Structure \\
           
        \textbf{RigL} \citep{Evci2020} & Any & Dynamic & Random to Heterogeneous & Magnitude & Full & High (sort/update) & Emergent & Training & Efficient Training \\
        {\scriptsize + SRigL} \citep{lasby2024dynamic} & & & & {\scriptsize Constant fan-in} & & & {\scriptsize Uniform (forced)} & & \\
        
        \textbf{SNIP} \citep{lee2018snip} & Any & Static & Dense to Sparse Uniform & Gradient $\times$ Weight & 1 Batch & Low & Problematic & One-shot & Prune-at-Init \\
    
        \textbf{GraSP} \citep{wang2020grasp} & Feedforward & Static & Dense to Sparse Uniform & Hessian-Gradient & 1 Batch & Medium & Uniform & One-shot & Gradient Flow \\
          
        \textbf{SynFlow} \citep{tanaka2020pruning} & Any & Static (Iterative) & Dense to Sparse Uniform & Gradient Flow & None & Medium (100 passes) & Uniform & Iterative & Data-free \\
                
        \textbf{Block Sparse} \citep{Gray2017} & Any & Static & Structured to Fixed Blocks & Configurable & N/A & Low & Optional & Hardware & Speedup \\
               
        \textbf{Mixture-of-Experts} \citep{Shazeer2017} & Any & Dynamic (Input) & Modular & Learned & Full & High (routing) & Module & Learned & Conditional Comp. \\

        \textbf{CHT/DNM} \citep{zhang2026brain, Cerretti2025dendritic} & Feedforward & Static/Dynamic & Topology Model & Magnitude (if dyn.) & None/Full & Low to High & Designed (parametric) & Dendritic / Fan-in & Initial Structure \\
         
        \textbf{EOI} \citep{Nowak2024eoi} & Any & Static & Orthogonal Sparse & N/A & None & Medium (constraints) & Uniform & Orthogonal & Initial Quality \\
         \midrule

        \textbf{PSN} (This Work) & Feedforward & Static & Profile to Heterogeneous & Profile Function & None & None & Designed & Mathematical Function & Inductive Bias \\

        \bottomrule
\end{tabularx}
\end{threeparttable}
\end{sidewaystable*}

Table~\ref{tab:sparse_comparison} provides comprehensive comparison of sparse training methods across key dimensions. The comparison reveals a fundamental divide between methods that discover structure (SET, RigL, LTH) and those that design it (PSN, Block Sparse). Discovery methods can potentially find task-specific optimal topologies but require expensive search and produce stochastic results. Design methods provide deterministic, analyzable structures but may miss task-specific adaptations. While these methods achieve impressive compression, they do not explore whether systematic connectivity variation provides inductive bias benefits. PSN bridges this by designing structures inspired by what discovery methods naturally find (heterogeneous fan-in organization), testing whether the emergent target structure provides benefits when used from initialization. The table also includes recent designed-topology methods (CHT/DNM, EOI) that share PSN's motivation of designing structure at initialization but differ in parameterization. CHT/DNM selects among discrete topological models with optional dynamic refinement, and EOI focuses on orthogonality rather than heterogeneity. Capacity distribution (controlled by profile function) determines how many total connections each output neuron receives, while input coverage (controlled by spreading pattern) determines which specific inputs connect to each output. Previous work in dynamic sparse training\citep{Evci2020, lasby2024dynamic} and block-structured methods\citep{Gray2017} conflates these dimensions, making it unclear whether benefits arise from having hubs (capacity heterogeneity) or from how hub connections are distributed across inputs (coverage pattern). PSN makes neuron-level connectivity variance an explicit architectural choice, contrasting with methods where heterogeneity emerges unpredictably from training dynamics.

Explicit parameterization through continuous profile functions in PSN enables systematic exploration of connectivity distribution space. In dynamic sparse training like RigL, the final connectivity pattern depends on initialization, hyperparameters, training data order, and stochastic optimization dynamics, making it difficult to isolate which structural properties drive performance benefits.\citep{Evci2020} Explicit normalization controls allow different profile shapes to be matched to target sparsity levels while preserving their relative structure, enabling fair comparisons across profile families at equal parameter counts. This addresses a critical experimental design issue in sparse network research, where methods are often compared at different effective parameter counts.\citep{Hoefler2021, Gale2019} Finally, PSN offers zero computational overhead compared to dynamic methods requiring topology updates, data-free structure definition unlike methods requiring training data or gradient information, and guaranteed avoidance of layer collapse through design rather than iterative correction. 

\section{Discussion}

This work maps the connectivity design space of sparse neural networks by varying fan-in distributions as a continuous architectural variable. Across four datasets and sparsity levels from 80\% to 99.9\%, PSN profiles spanning fan-in CV from 0 (uniform random) to 2.5 (highly heterogeneous) produce no statistically significant accuracy differences. This holds across eight named profiles, lognormal and powerlaw families with continuously varied CV, and two spreading patterns. The random baseline, which can be understood as PSN with CV\,=\,0, matches or exceeds all structured profiles. At the same time, dynamic sparse training converges to a characteristic fan-in topology regardless of initialisation, and matching that topology at initialisation consistently improves final accuracy.

On these tasks, networks with sufficient capacity are insensitive to how that capacity is distributed across neurons. The multi-peak interpolation experiment makes this explicit. Fan-in CV varies continuously from 3.2 to 0 with no measurable effect on accuracy, even as the underlying connectivity changes from extreme concentration (a few hub neurons receiving all connections) to complete uniformity.

Random projection theory offers an intuitive, though not causal, explanation. A sparse neural layer computes weighted sums over subsets of inputs, collectively forming a random linear mapping from input space to a lower-dimensional representation. Results from high-dimensional geometry show that such mappings can approximately preserve pairwise distances between points in a dataset when the embedding dimension scales logarithmically with the number of points.\citep{JohnsonLindenstrauss1984} This guarantee concerns geometric structure rather than classification performance directly, but it suggests that randomly initialized sparse connectivity can retain sufficient information for downstream learning. Estimates of MNIST's intrinsic dimensionality typically place it in the low tens. At 90\% sparsity, each neuron receives approximately 78 random inputs, well above this scale, suggesting that random sparse connectivity is plausibly informationally sufficient for representing the data. In such regimes, structured connectivity profiles are unlikely to provide large gains because the representation already captures the relevant geometric structure. This reasoning predicts that performance should degrade when mean fan-in approaches the intrinsic dimensionality, which occurs near 98\% sparsity for 784-input datasets (fan-in $\approx$ 16) and around 80\% for Forest Cover (fan-in $\approx$ 11). Consistent with this prediction, the sharpest accuracy drops in our data occur in precisely these regimes. When random features are sufficient, static connectivity structure may play a limited role and weight optimization dominates performance. When capacity becomes limiting, at higher sparsity, on more complex tasks, or at larger scale, the specific pattern of connections becomes increasingly important, and adaptive methods that select task-relevant connections often outperform static ones.

RigL converges to a characteristic fan-in CCV that depends on architecture and sparsity but not on the task or initialisation. Lognormal initialization matched to this equilibrium CCV outperforms ERK across all datasets, with the advantage growing from negligible on MNIST to +0.16\% on Fashion-MNIST ($p = 0.036$, $d = 1.07$), +0.43\% on EMNIST, and +0.49\% on Forest Cover. This pattern of increasing advantage with task difficulty is consistent with the broader sparse training literature, where the gap between static and dynamic methods grows from $\sim$1\% on MNIST\citep{Mocanu2018} to 1.4--1.7\% on CIFAR-10 and 4.3--7.0\% on ImageNet.\citep{Evci2020} Starting at the equilibrium topology eliminates early topology search, allowing the optimiser to invest gradient steps in weight refinement from the outset. When dynamic sparse training converges to a predictable structural endpoint, that endpoint can be computed analytically and used as initialisation, reducing the computational overhead of topology discovery without sacrificing the benefits of heterogeneous connectivity.

All experiments use fully-connected architectures on classification tasks where dense baselines already achieve high accuracy. Whether the RigL initialisation advantage persists in convolutional or transformer architectures, at larger scales, or on tasks where capacity is genuinely limiting remains an open question. The effect sizes, while statistically significant, are small in absolute terms. Our implementation applies sparse masks to dense weight matrices ($W_{\text{sparse}} = M \odot W$), so no actual computational savings are realised. True sparse matrix kernels would be needed to translate sparsity into wall-clock speedups. The static results may not generalise to regimes where network capacity is scarce relative to task complexity. On the benchmarks tested here, random sparse features are sufficient (as the J-L analysis suggests), so connectivity structure is irrelevant to accuracy. On harder tasks such as ImageNet, language modelling, or settings where dense baselines themselves struggle, capacity becomes limiting, which connections exist begins to matter, and structured initialisation may provide larger advantages. This is precisely the regime where the literature shows the largest gaps between static and dynamic sparse methods (4.3--7.0\% on ImageNet\citep{Evci2020}), and where PSN-style equilibrium initialisation warrants further investigation.

\section{Conclusions}

We introduced PSN, a framework for mapping the connectivity design space of sparse neural networks through continuous profile functions. Across four benchmarks, eight profile shapes, fan-in CVs from 0 to 2.5, and sparsity levels from 80\% to 99.9\%, static connectivity structure does not significantly affect accuracy at matched parameter counts. The degree of heterogeneity is irrelevant when hub placement is arbitrary rather than task-aligned.

Dynamic sparse training converges to a characteristic fan-in distribution regardless of starting point, and this equilibrium topology can be computed analytically from the network architecture and target sparsity. Lognormal profiles matched to this equilibrium consistently outperformed standard ERK initialisation, with advantages increasing on harder tasks (+0.49\% on Forest Cover, +0.43\% on EMNIST). Understanding the structural endpoints of dynamic sparse training provides a principled basis for initialisation design, a direction that warrants further investigation on larger-scale problems where the gap between static and dynamic sparse training is known to be substantial.

\bibliographystyle{unsrt}
\bibliography{bibliography}

\end{document}